\documentclass[letterpaper]{article} % DO NOT CHANGE THIS
\usepackage{aaai25}  % DO NOT CHANGE THIS
\usepackage{times}  % DO NOT CHANGE THIS
\usepackage{helvet}  % DO NOT CHANGE THIS
\usepackage{courier}  % DO NOT CHANGE THIS
\usepackage[hyphens]{url}  % DO NOT CHANGE THIS
\usepackage{graphicx} % DO NOT CHANGE THIS
\usepackage{natbib}  % DO NOT CHANGE THIS AND DO NOT ADD ANY OPTIONS TO IT
\usepackage{caption} % DO NOT CHANGE THIS AND DO NOT ADD ANY OPTIONS TO IT
\usepackage{algorithm}
\usepackage{algorithmic}
\usepackage{multirow}
\usepackage{booktabs}
\usepackage{colortbl}
\usepackage{amssymb}
\usepackage[utf8]{inputenc}
\usepackage{newunicodechar}
\newunicodechar{ℝ}{\mathbb{R}}
\usepackage{amsmath}
\usepackage{lipsum}
\usepackage{bbding}
\usepackage{pifont}
\usepackage{xcolor}
\usepackage{mathrsfs}
\usepackage{newfloat}
\usepackage{listings}
\DeclareCaptionStyle{ruled}{labelfont=normalfont,labelsep=colon,strut=off} % DO NOT CHANGE THIS
\floatstyle{ruled}
\newfloat{listing}{tb}{lst}{}
\floatname{listing}{Listing}
\title{MUSE: Mamba is Efficient Multi-scale Learner for Text-video Retrieval}
\author{
    Haoran Tang\textsuperscript{\rm 1,2}, 
    Meng Cao\textsuperscript{\rm 1},
    Jinfa Huang\textsuperscript{\rm 1},
    Ruyang Liu\textsuperscript{\rm 1,2},
    Peng Jin\textsuperscript{\rm 1,2},
    Ge Li\textsuperscript{\rm 1}\thanks{Corresponding Author},
    Xiaodan Liang\textsuperscript{\rm 3}
}
\affiliations{
    \textsuperscript{\rm 1} Guangdong Provincial Key Laboratory of Ultra High Definition Immersive Media Technology,\\ Shenzhen Graduate School, Peking University\\
    \textsuperscript{\rm 2} Peng Cheng Laboratory\\
    \textsuperscript{\rm 3} Sun Yat-sen University
}

\begin{document}
\maketitle
\begin{abstract}
Text-Video Retrieval (TVR) aims to align and associate relevant video content with corresponding natural language queries. Most existing TVR methods are based on large-scale pre-trained vision-language models (e.g., CLIP). However, due to CLIP's inherent \emph{plain} structure, few TVR methods explore the multi-scale representations which offer richer contextual information for a more thorough understanding. To this end, we propose MUSE, a multi-scale mamba with linear computational complexity for efficient cross-resolution modeling. Specifically, the multi-scale representations are generated by applying a feature pyramid on the last single-scale feature map. Then, we employ the Mamba structure as an efficient multi-scale learner to \emph{jointly} learn scale-wise representations. Furthermore, we conduct comprehensive studies to investigate different model structures and designs. Extensive results on three popular benchmarks have validated the superiority of MUSE.
\end{abstract}
\begin{links}
    \link{Code}{https://github.com/hrtang22/MUSE}
    \link{Extended version}{https://arxiv.org/abs/2408.10575}
\end{links}
\section{Introduction}
Text-Video Retrieval (TVR) \cite{gabeur2020multi,gorti2022x,he2021improving,lei2021less,luo2022clip4clip,ma2022x,wang2022disentangled} is a fundamental task in multimodal research. Its objective is to locate the most relevant video content within a repository in response to a text query and vice versa. 

Based on large-scale image-text pre-trained model CLIP \cite{radford2021learning}, most current TVR methods focus on transferring CLIP to the video-text domain. To achieve fine-grained representations, mainstream methods capture cross-modal alignment at different granularities, including video-sentence \cite{ma2022x}, frame-sentence \cite{gorti2022x,ma2022x} or even patch-word \cite{wang2023unified} levels. 

\begin{figure*}[t]

\centering
    \captionsetup{type=figure}
    \includegraphics[width=1.0\textwidth]{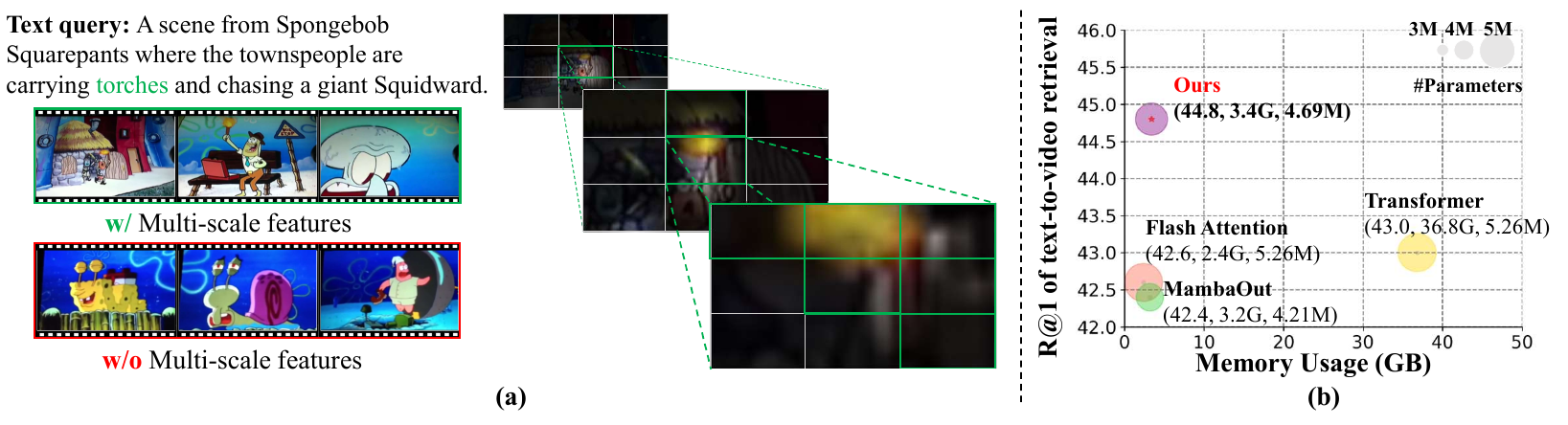}
    \captionof{figure}{\textbf{(a) Illustration of multi-scale features.} Giving the text query, the model without multi-scale features retrieves the relevant but incorrect video because the small but crucial object \emph{``torches"} can not be identified by only using frame-level feature representation (e.g., \texttt{[CLS]} tokens). We visualize the token similarity of the word \emph{``torches"} and our extracted multi-scale features by organizing the attention map in a feature pyramid style from resolution low to high. Our model aggregates patches of the object \emph{``torches"} that have a green boundary from multiple granularities to finally build a correlation between word \emph{``torches"} and its visual entity in the video; \textbf{(b) Efficiency-performance comparisons.} The horizontal axis reflects memory usage, and the vertical is the R@1 metric of text-to-video retrieval on the MSR-VTT dataset. Marker sizes are proportional to the number of tunable parameters. Memory and parameters are calculated only on video learners without adding the backbone.\\}
    % %\vspace{-0.7cm}
    \label{fig:motivation}
\end{figure*}

However, CLIP is inherently a \emph{plain} structure with the identical token length for all the layers. Therefore, these methods ignore the exploration of representations of different scales, which provides more valuable contextual information for comprehensive understanding. For example, in Figure \ref{fig:motivation}(a), the textual query aims to retrieve a video where ``people are carrying \texttt{torches} and chasing a giant squidward". As shown, the most discriminative object \texttt{torch} is not highlighted in the original resolution\footnote{We use ``scale" for feature-level representation and ``resolution" for pixel-level raw image.}. The loss of such detailed yet important information leads to incorrect retrieval results. In contrast, we can see that the torch region is correctly highlighted when focusing on higher-resolution representations. Therefore, information hidden in higher resolution should be considered for TVR.

In light of this, two issues naturally arise. 1) how to \emph{generate} multi-scale representations? Since the vanilla CLIP architecture is non-hierarchical, it maintains a single-scale feature map. Following the spirit of ``fewer inductive biases" proposed in \cite{li2022exploring}, we build a feature pyramid from the last single-scale feature map via convolution or pooling operations. Compared to the ConvNet-based methods, \emph{e.g.}, Swin Transformers \cite{liu2021swin}, such design does not require the introduction of additional modules and is, therefore, more efficient; 2) how to efficiently model \emph{cross-resolution} correlations? One intuitive idea is to jointly model various resolutions in a holistic manner, \emph{i.e.}, flattening resolution-wise representations and modeling the comprehensive correlations with the widely-used attention mechanism \cite{vaswani2017attention}. This strategy inevitably introduces a huge amount of computation, which is quadratically correlated with the length of the sequence (e.g., Transformer requires 36.8GB GPU memory when the input frame is 12, as shown in Figure \ref{fig:motivation}(b)). To this end, we argue that \textbf{M}amba is an efficient m\textbf{U}lti-\textbf{S}cal\textbf{E} learner (dubbed as \textbf{MUSE}) for text-video retrieval. Specifically, MUSE is proposed with linear computational complexity for efficient cross-resolution modeling. Through extensive experiments, we can conclude that Mamba-like structures are efficient cross-resolution context learners, which leads to superior performance compared to the Transformer-based methods. As shown in Figure \ref{fig:motivation}(b), our MUSE achieves state-of-the-art performance on the MSR-VTT dataset and a relatively small memory footprint and tunable parameters.
 
Since the community has few empirical experiences modeling multi-scale correlations in a linear complexity, we conduct extensive exploratory studies to find the optimal training architecture. We explore the following aspects: 1) \textbf{Plug-and-play manner}. The proposed MUSE is model-agnostic and can be compatible with existing TVR methods; 2) \textbf{Correlation modeling strategies}: We experiments popular architectures including MambaOut \cite{yu2024mambaout}, FlashAttention \cite{dao2022flashattention}, and Mamba \cite{zhu2024vision}; 3) \textbf{Scan strategies}: We experiment with existing sequence scan manners to find the optimal design; 4) \textbf{Scale combination manners}: Obtaining the representations in various scales, how to combine and arrange these representations is worth exploring. We hope our extensive explorations can shed light on effective and efficient linear attention modeling in multi-scale scenarios.
 
To conclude, the main contributions of this work are:

\begin{itemize}
    \item We propose MUSE to explore the multi-scale representations for TVR, which are generated by applying a feature pyramid on the last single-scale feature map.

    \item We experiment with both Transformer and popular linear-attention architectures for joint resolution modeling and argue that Mamba is an efficient multi-scale learner for TVR.
	
    \item Extensive experiments show that our proposed MUSE achieves state-of-the-art performance on MSR-VTT, DiDeMo, and ActivityNet benchmarks.  
\end{itemize}
\section{Related Works}\label{related_works}

\begin{figure*}[t]
\centering
\includegraphics[width=0.9\textwidth]{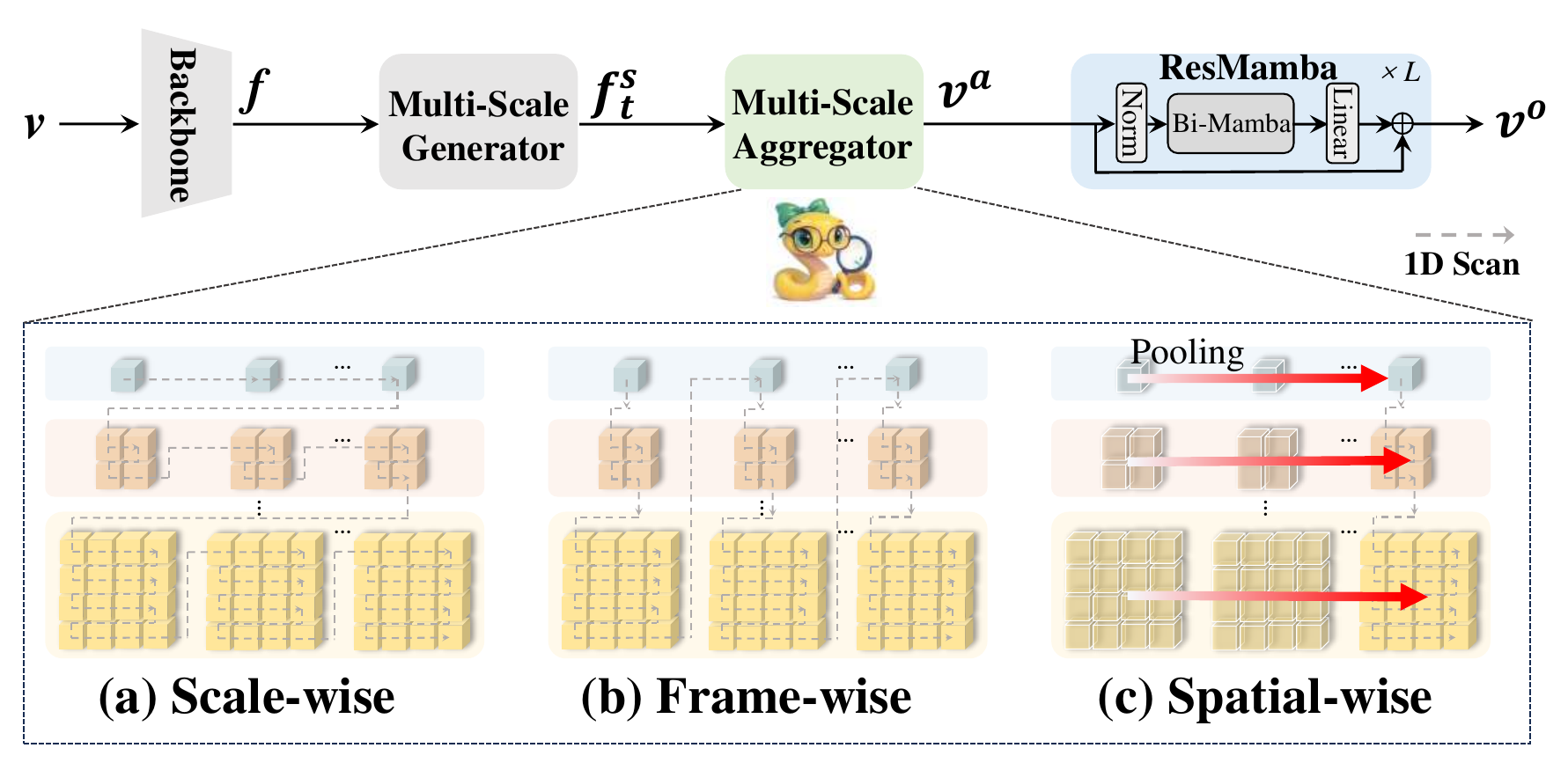}
%\includegraphics[width=1.0\textwidth]{figs/pipelineV2.pdf}
%\vspace{-0.3cm}
\caption{\textbf{Illustration of MUSE.} Our proposed method consists of three modules applied after video backbones. The generation module generates multi-scale video features based on single-scale visual output. Then, for the aggregation module, we test three different aggregation manners to aggregate multi-scale features into a 1D sequence. Finally, we design a residual architecture with Mamba to capture crucial video information from different granularities.}
%\vspace{-0.3cm}
\label{fig:pipeline}
\end{figure*}

\noindent \textbf{Text-Video Retrieval.} TVR \cite{yu2018joint,gabeur2020multi,lei2020tvr,gorti2022x,he2021improving,lei2021less,luo2022clip4clip,ma2022x,wang2022disentangled,cao2024rap,jiang2022tencent, jin2023text,jin2022expectation,jin2023diffusionret,wu2024uncertainty} is a pivotal task for video cross-modal learning, which has widespread application in video understanding \cite{liu2024st,zhang2021cola,zhang2022unsupervised,li2023g2l,cao2022deep,cao2022correspondence,li2023generating,cao2023iterative,cao2022correspondence} and multi-modal interactions \cite{cao2021pursuit,cao2022locvtp,liu2023qilin,ye2023qilin,ji2023video,luo2024textual,liu2024bt,yang2021rr,yang2023concept,li2024exploiting}. With the advancement of image-text pretraining, recent works \cite{luo2022clip4clip, ma2022x, tu20222, gorti2022x, jiang2022tencent, jin2023text} resort to image-text pretraining model CLIP \cite{radford2021learning} and focus on image-to-video transferring fine-tuning. The primary work CLIP4clip \cite{luo2022clip4clip} investigates three kinds of temporal aggregation manners, which enlightens the follow-up works. To achieve fine-grained representations, X-CLIP \cite{ma2022x} explores cross-grained contrastive learning, including video-sentence, video-word, frame-word and frame-sentence. Hunyuan\_tvr \cite{jiang2022tencent} divides the video-language interaction into frame-word, clip-phrase, and video-sentence granularities. DiCoSA \cite{jin2023text} further improves the fine-grained alignments by disentangling video into visual concepts. However, these methods neglect the cross-resolution relationships, which offer another perspective over the resolution-wise feature correlations. Our proposed MUSE bridges this gap by presenting scale-aware representations.

\noindent \textbf{Multi-scale Video Modeling.} Recent progress in image detection and segmentation finds that simply using the ViT \cite{dosovitskiy2020image} output feature is insufficient for fine-grained image understanding. Thus, ViTDet \cite{li2022exploring} has explored the plain vision transformer architecture to build a feature pyramid from ViT outputs, which achieves progress on dense image prediction. ViT-Adapter \cite{chen2022vision} trains an additional visual adapter to obtain multi-scale representations which considers both tasks prior and the input images. For video understanding, SlowFast \cite{feichtenhofer2019slowfast} uses the temporal-wise multi-scale branches with both low and high frame rates. MS-TCT \cite{dai2022ms} proposes a temporal scale mixer module to effectively fuses multi-scale features. These methods fails to jointly model cross-scale features in a holistic manner. Therefore, we leverage Mamba \cite{gu2023mamba} of linear complexity as multi-scale video learner and design different manners for feature aggregation.

\noindent \textbf{Mamba for Video.} Based on the success of Mamba in language modeling, Vim \cite{zhu2024vision} and VMamba \cite{liu2024vmambavisualstatespace} has pioneered Mamba architecture in vision by designing a bidirectional State Space Model and 2D selective scan manners for image recognition. The follow-up methods \cite{yang2024plainmamba,huang2024localmamba, pei2024efficientvmamba} explore different architecture design and selective scan manners. Also, some works \cite{shen2024gamba, ma2024u, liu2024point, chen2024rsmamba} have extended the success of Mamba to areas such as medical image processing, 3D reconstruction, point cloud understanding, etc. For video understanding, VideoMamba \cite{li2024videomamba} first trains a video foundation model with Mamba backbone using the paradigm of unmasked teacher \cite{li2023unmasked} and shows its efficiency compared with the transformer counterpart. Video mamba suite \cite{chen2024video} explores Mamba's effectiveness in the video downstream tasks by replacing attention blocks with its DBM blocks. In this work, we built a Mamba learner with a simple gated structure based on the Bidirectional Mamba \cite{zhu2024vision}. We find that Mamba is an efficient multi-scale video learner that surpasses Transformer and other linear attention methods in text-video retrieval.
\definecolor{babyblue}{rgb}{0.54, 0.81, 0.94}
\definecolor{lightblue}{rgb}{0.68, 0.85, 0.9}
\definecolor{lightcornflowerblue}{rgb}{0.6, 0.81, 0.93}
\definecolor{lightgray}{rgb}{0.83, 0.83, 0.83}
\definecolor{aliceblue}{rgb}{0.94, 0.97, 1.0}
\definecolor{deeppink}{RGB}{255,20,147}
\definecolor{mygray}{gray}{0.95}
\section{Methodology} \label{method}

In this section, we first introduce the architecture of standard text-video retrieval methods and how to extract multi-scale video features. Then, we thoroughly present our proposed method MUSE which consists of three components: Multi-scale generator, Multi-scale aggregator and ResMamba.
\subsection{Overview}
\noindent \textbf{Feature extraction.} Given a video $\boldsymbol{v} \in \mathcal{V}$ and the corresponding query $\boldsymbol{t} \in \mathcal{T}$, we utilize CLIP \cite{radford2021learning} with ViT \cite{dosovitskiy2020image} to extract features. For the textual branch, we select the \texttt{[EOT]} token as text representations following \cite{radford2021learning}. For the visual branch, former methods (e.g. CLIP4clip \cite{luo2022clip4clip}) regard the frame-wise \texttt{[CLS]} token as the video-level representations $\boldsymbol{f}_{cls} \in \mathbb{R}^{T \times C}$. In contrast, to extract fine-grained video information, we utilize all visual tokens as video-level representations $\boldsymbol{f} \in \mathbb{R}^{T \times N \times C}$, where $T$ is the frame number and $N$ denotes the number of visual tokens.\\
\noindent \textbf{Multi-scale feature generation.} As shown in Figure \ref{fig:pipeline}, based on visual features $\boldsymbol{f}$, we aim to extract multi-scale video features $\boldsymbol{f}_{t}^{s}$, which denotes the feature representations for $t$-th frame at the scale $\boldsymbol{s}_i$, $s\in[1,S]$. We follow ViTDet \cite{li2022exploring} by applying convolution or pooling operations on $\boldsymbol{f}$.
\begin{equation}
    \boldsymbol{f}_{t}^{s} = \operatorname{Pool} \big(\operatorname{Conv} (\boldsymbol{f})\big).
\end{equation}
\noindent \textbf{Optimization.} We follow \cite{radford2021learning} by using cross-entropy loss for optimization.
\begin{equation}
\mathcal{L}=-\log \frac{ \exp (\boldsymbol{v} {\cdot} \boldsymbol{t} / \tau)}{\sum_{\boldsymbol{t}^{-}} \exp \left(\boldsymbol{v} {\cdot} \boldsymbol{t}^{-} / \tau\right)}, 
\end{equation}
\noindent where $\boldsymbol{t}^{-}$ is the unmatched language query.
\subsection{Multi-scale Feature Aggregation}
To conduct joint cross-resolution feature aggregations, we design three different methods: scale-wise, frame-wise, and spatial-wise.\\
\noindent\textbf{Scale-wise}. We first aggregate the video features from the same scale by temporal order, then rearrange the tokens from the same scale as a 1D sequence as shown in Figure \ref{fig:pipeline}(a). Finally, we concatenate tokens following scale orders from low resolution to high resolution. After aggregation, the video feature can be formulated as:
\begin{equation}
    \begin{aligned}
        \boldsymbol{f}^{s} &= \{\boldsymbol{f}_{1}^{s}, \boldsymbol{f}_{2}^{s}, \cdots, \boldsymbol{f}_{T}^{s}\} \\
        \boldsymbol{v}^{a} &= \{\boldsymbol{f}^{{s}}\}_{s=1}^{S}.
    \end{aligned}
\end{equation}
\noindent \textbf{Frame-wise}. Different from the scale-wise manner, we first aggregate the video tokens from the same video frame and then rearrange them as a 1D sequence following scale order as shown in Figure \ref{fig:pipeline}(b). Finally, the video tokens are concatenated in frame order. In this manner, the aggregated video feature can be formulated as follows:
\begin{equation}
    \begin{aligned}
        \boldsymbol{f}_{t} &= \{\boldsymbol{f}_{t}^{1}, \boldsymbol{f}_{t}^{2}, \cdots, \boldsymbol{f}_{t}^{S}\} \\
        \boldsymbol{v}^{a} &= \{\boldsymbol{f}_{{t}}\}_{t=1}^{T}.
    \end{aligned}
\end{equation}
\noindent\textbf{Spatial-wise}. In this manner, the aggregate order is the same as frame-wise for each frame. Differently, we first pool the video tokens by temporal dimension to aggregate temporal information and only keep spatial information. Then, we rearrange the tokens in a frame-wise manner. This manner is spatial-wise as it only focuses on the spatial dimension. Details are shown in Figure \ref{fig:pipeline}(c), and the final video feature can be formulated as follows:
\begin{equation}
   \begin{aligned}
    \boldsymbol{f}^{s} &= \operatorname{meanpool} (\boldsymbol{f}_{1}^{s}, \boldsymbol{f}_{2}^{s}, \cdots, \boldsymbol{f}_{T}^{s}) \\
    \boldsymbol{v}^{a} &= \{\boldsymbol{f}^{{s}}\}_{s=1}^{S}.
    \end{aligned} 
\end{equation}
In practice, We find that rearranging multi-scale video features as a 1D sequence in a scale-wise manner achieves the best performance as shown in Table \ref{tab:ablate_agg}. In Section 4, we will examine different scan methods and explain why scale-wise is effective. In the following, we will explain how we design the feature aggregation variants.
\subsection{Mamba As Video Learner}
To modify Mamba as an effective multi-scale video learner, we design a residual network following TimeSformer \cite{bertasius2021space} that can be noted as: 
\begin{equation}
       \boldsymbol{v}^{o} = \operatorname{ResMamba}(\boldsymbol{v}^{a}).
    \label{eq:9}
\end{equation}
Specifically, with experiments we find that gated residual architecture works best for multi-scale video learning. In practice, we leverage a single \emph{Linear} layer with zero initialization after Mamba block as gated function $\mathcal{G}(.)$ and the learning process can be formulated as:
\begin{equation}
    \begin{aligned}
       \boldsymbol{h}_{l} &=\mathbf{A}\boldsymbol{h}_{l-1} + \mathbf{B}\boldsymbol{v}^{a}_{l}\\
       \boldsymbol{y}_{l} &=\mathbf{C}\boldsymbol{h}_{l}\\
       \boldsymbol{v}^{a}_{l+1} &= \mathcal{G}(\operatorname{Norm}(\boldsymbol{y}_{l})) +  \boldsymbol{v}^{a}_{l},
    \end{aligned}
\end{equation}
where $l$ denotes the $l_{th}$ layer in $\boldsymbol{\emph{L}}$ Mamba layers. For Mamba block, $\mathbf{A}\in \mathbb{R}^{N\times N}$ is the evolution parameter, $\mathbf{B}\in \mathbb{R}^{N\times 1}$ and $\mathbf{C}\in \mathbb{R}^{1\times N}$ are the projection parameters.
\definecolor{babyblue}{rgb}{0.54, 0.81, 0.94}
\definecolor{lightblue}{rgb}{0.68, 0.85, 0.9}
\definecolor{lightcornflowerblue}{rgb}{0.6, 0.81, 0.93}
\definecolor{lightgray}{rgb}{0.83, 0.83, 0.83}
\definecolor{aliceblue}{rgb}{0.94, 0.97, 1.0}
\definecolor{deeppink}{RGB}{255,20,147}
\definecolor{mygray}{gray}{0.95}
\section{Experiment}

\subsection{Experimental Settings}
\noindent \textbf{Implementation Details.}
We set the input frame
length to 12, 64, 64 and the caption token length
to 32, 64, and 64 for MSR-VTT, DiDeMo, and ActivityNet, respectively. For fine-tuning, we keep the training hyperparameters and settings of the base model unchanged and train MUSE with a learning rate of 10 times higher (e.g., 1e-4 for CLIP4clip and 1e-3 for MUSE). The Layer number of ResMamba is set to 4, and the scale selected is \{1, 3, 7, 14\}. All experiments were
carried out on 8 NVIDIA A100 GPUs.\\
\noindent \textbf{Evaluation Metrics.}
For evaluation, we test the performance with standard retrieval metrics following CLIP4clip \cite{luo2022clip4clip}, which includes recall at rank $K$ (R@$K$, higher is better), median rank (MdR, lower is better) and mean rank (MnR, lower is better). R@$K$ defines the recall percentage of samples whose correct answer is found in the top-$K$ retrieved results. We set $K$ to $\{1, 5, 10\}$ following CLIP4clip \cite{luo2022clip4clip}. MdR is defined as the median of the ground-truth results rank in the result ranking list, while MnR is defined as the mean rank of all the correct results.\\
\noindent \textbf{Datasets.}
To validate the effectiveness of our proposed model(MUSE), we test our model on three benchmarked datasets:  \textbf{MSR-VTT} \cite{xu2016msr} contains 10,000 YouTube videos, and each video is associated with 20 textual descriptions. We follow the 1k-A split \cite{yu2018joint} where 9,000 videos are used for training and 1,000 videos for testing.
\textbf{ActivityNet} \cite{krishna2017dense} comprises 20,000 untrimmed videos of complex human activities with an average duration of two minutes. We report results on the ``val1'' split (including 10,009 training videos and 4,917 testing videos) following \cite{gabeur2020multi}.
\textbf{DiDemo} \cite{anne2017localizing} consists of 10,464 unedited, personal videos in diverse visual settings annotated with 40,543 text descriptions. We follow the training and evaluation protocol in \cite{luo2022clip4clip}.

\begin{table*}[t]
\centering
\renewcommand\arraystretch{1.1}
\begin{tabular}{l|lll|lll}
\toprule
\multirow{2}{*}{Methods} &  \multicolumn{3}{c|}{\textbf{Text-\textgreater{}Video}}        & \multicolumn{3}{c}{\textbf{Video-\textgreater{}Text}}  \\  
& R@1$\uparrow$ & R@5$\uparrow$ & R@10$\uparrow$
& R@1$\uparrow$ & R@5$\uparrow$ & R@10$\uparrow$ \\
\midrule
CLIP4Clip$^{\dag}$ \cite{luo2022clip4clip}  & 42.6 & 70.8 & 79.9  & 43.9 & 70.0 & 81.4  \\
\rowcolor{aliceblue}\bf + MUSE (Ours) & \bf 44.8 (+2.2) & \bf 71.6 (+0.8) & \bf 82.1 (+2.2)   & \bf44.9 (+1.0)  & \bf 70.8 (+0.8)  & \bf 82.2 (+0.8)  \\
\midrule
EMCL-Net$^{\dag}$ \cite{jin2022expectation}  & 47.1 & 72.7 & 82.3   & 44.4 & 72.6 & 82.6   \\
\rowcolor{aliceblue}\bf + MUSE (Ours) & \bf 48.8 (+1.7)  & \bf 74.1 (+1.4)  & \bf 83.4 (+1.1) & \bf 47.4 (+3.0)  & \bf 75.8 (+3.2)  & \bf 82.9 (+0.3)          \\
\midrule
STAN$^{\dag}$ \cite{liu2023revisiting}  & 46.2 & 72.6 & 81.1 & 44.5 & 71.9 & 81.7   \\
\rowcolor{aliceblue}\bf + MUSE (Ours) & \bf 47.3 (+1.1)  & \bf 73.1 (+0.5)  & \bf 82.2 (+1.1)  & \bf 45.5 (+1.0)  & \bf 73.1 (+1.4) & \bf 81.8 (+0.1)  \\
\midrule
T-MASS$^{\dag}$ \cite{wang2024text}  & 50.0 & 75.3 & 84.2  & 46.0 & 77.1 & 86.2   \\
\rowcolor{aliceblue}\bf + MUSE (Ours) & \bf 50.9 (+0.9)  & \bf 76.7 (+1.5)  & \bf 85.6 (+1.4)  &  \bf 49.7 (+3.7)  & \bf 77.8 (+0.7)  & \bf 86.5 (+0.3)  \\
\bottomrule
\end{tabular}
\caption{Plug-and-play experiments on MSR-VTT. We compare the text-to-video retrieval results before and after adding our proposed method, \textbf{MUSE}, on four baseline models. $\dagger$ denotes our reproduction of the method.}
\label{tab:ablate_plug_and_play}
\end{table*}

\begin{table*}[t]
\renewcommand\arraystretch{1.1}
\centering
\resizebox{\linewidth}{!}
{
\begin{tabular}{l|ccccc|ccccc}
\toprule%[1.5pt]
  & \multicolumn{5}{c|}{Text → Video} & \multicolumn{5}{c}{Video → Text} \\
 Methods & R@1↑ & R@5↑ & R@10↑ & MdR↓ & MnR↓ & R@1↑ & R@5↑ & R@10↑ & MdR↓ & MnR↓ \\ 
\midrule
CLIP4Clip\cite{luo2022clip4clip} & 44.5 &  71.4 & 81.6 & 2.0 & 15.3 & 42.7 & 70.9 & 80.6 & 2.0 & 11.6 \\
X-Pool\cite{gorti2022x} & 46.9 &  72.8 & 82.2 & 2.0 & 14.3 & 44.4 & 73.3 & 84.0 & 2.0 & 9.0 \\
STAN\cite{liu2023revisiting} & 46.9 &  72.8 & 82.8 & 2.0 & - & - & - & - & - & - \\ 
EMCL-Net\cite{jin2022expectation} & 46.8 &  73.1 & 83.1 & 2.0 & - & 46.5 & 73.5 & 83.5 & 2.0 & - \\ 
HBI\cite{jin2023video} & 48.6 & 74.6 & 83.4 & 2.0 & 12.0 & 46.8 & 74.3 & 84.3 & 2.0 & 8.9 \\ 
DiffusionRet\cite{jin2023diffusionret} & 49.0 & 75.2 & 82.7 & 2.0 & 12.1 & 47.7 & 73.8 & 84.5 & 2.0 & 8.8 \\ 
T-MASS\cite{wang2024text} & 50.2 &  75.3 & 85.1 & 1.0 & 11.9 & 47.7 & \bf 78.0 & 86.3 & 2.0 & 8.0 \\ 
\rowcolor{aliceblue}\textbf{MUSE (Ours)} & \textbf{50.9} &  \textbf{76.7} & \textbf{85.6} & \textbf{1.0} & \textbf{10.9} & \textbf{49.7} & \underline{77.8} & \textbf{86.5} & \textbf{2.0} & \textbf{7.4} \\ 
\bottomrule %[1.5pt]
\end{tabular}
}
\caption{Comparisons with state-of-the-art methods on MSR-VTT dataset. Models are tested with CLIP-ViT-B/32\cite{radford2021learning}. The best performance is in \textbf{bold} and the second best is \underline{underlined}. 
}
\label{tab:msrvtt_SOTA}
\end{table*}

\begin{table*}[t]
\renewcommand\arraystretch{1.1}
%\vspace{-5pt}
\resizebox{\linewidth}{!}{
\begin{tabular}{llcccccccccc}
\toprule%[1.5pt]
\multicolumn{1}{c}{} & \multicolumn{5}{c}{DiDeMo} & \multicolumn{5}{c}{ActivityNet} \\ 
\multicolumn{1}{l}{Methods} & \multicolumn{1}{c}{R@1↑} & \multicolumn{1}{c}{R@5↑} & \multicolumn{1}{c}{R@10↑} & \multicolumn{1}{c}{MdR↓} & \multicolumn{1}{c}{MnR↓} & \multicolumn{1}{c}{R@1↑} & \multicolumn{1}{c}{R@5↑} & \multicolumn{1}{c}{R@10↑} & \multicolumn{1}{c}{MdR↓} & \multicolumn{1}{c}{MnR↓} \\ 
\midrule
\multicolumn{1}{l|} {CLIP4Clip \cite{luo2022clip4clip}} &  43.4 & 70.2 & 80.6 & 2.0 & \multicolumn{1}{l|} {{17.5}} &  {40.5} & {72.4} & - & 2.0 &{7.4} \\ 
\multicolumn{1}{l|}{X-CLIP\cite{ma2022x}} &  45.2 & 74.0 & - & - & \multicolumn{1}{l|}{14.6} & {44.3} & {74.1} & - & - &{7.9} \\ 
\multicolumn{1}{l|}{HBI\cite{jin2023video}} &  46.9 & 74.9 & 82.7 & 2.0 & \multicolumn{1}{l|} {{12.1}} & {42.2} & {73.0} & 84.6 & 2.0 &{6.6} \\
\multicolumn{1}{l|}{DiCoSA\cite{jin2023text}} &  45.7 & 74.6 & 83.5 & 2.0 & \multicolumn{1}{l|} {11.7} & {42.1} & {73.6} & 84.6 & 2.0 &{6.8} \\
\multicolumn{1}{l|}{DiffusionRet\cite{jin2023diffusionret}} &  46.7 & 74.7 & 82.7 & 2.0 & \multicolumn{1}{l|} {{14.3}} & 45.8 & 75.6 & 86.3 & 2.0 &{6.5} \\
\multicolumn{1}{l|}{T-MASS\cite{wang2024text}} & 50.9 & 77.2 & 85.3 & 1.0 & \multicolumn{1}{l|} {{12.1}} & - & - & - & - & - \\
\rowcolor{aliceblue} \multicolumn{1}{l|}{\textbf{MUSE(Ours)}} & \bf 51.5 & \bf 77.7 & \bf 86.0 & \bf 1.0 & \multicolumn{1}{l|} {\bf 11.3} & \bf 46.2 & \bf 76.9 & \bf 86.8 & \bf 2.0 & \bf5.8 \\
\bottomrule%[1.5pt]
\end{tabular}
%\vspace{-0.2cm}
}
\caption{Comparisons with state-of-the-art methods on DiDeMo and ActivityNet Datasets.}
\label{tab:other_SOTA}
\end{table*}

\subsection{Performance Comparison}
In this section, we validate the effectiveness and generalizability of our proposed method MUSE. Table \ref{tab:ablate_plug_and_play} shows that our method can be applied to modern text-video retrieval models as a plug-and-play module. In Table \ref{tab:msrvtt_SOTA} and Table \ref{tab:other_SOTA}, we compare our performance with state-of-the-art (SOTA) methods, and we find that directly appending MUSE to the former SOTA methods can achieve new SOTA performance on three widely used benchmarks.\\
\noindent \textbf{Plug-and-play.} 
To validate the generalizability of our proposed method, we test MUSE on four mainstream TVR baselines, including CLIP4clip \cite{luo2022clip4clip}, EMCL-Net \cite{jin2022expectation}, STAN \cite{liu2023revisiting}, and T-MASS \cite{wang2024text}. Specifically, we add MUSE as a video aggregator after video feature extraction of the baseline models, without any special modification except adjusting the learning rate of MUSE. As shown in Table \ref{tab:ablate_plug_and_play}, our proposed method outperforms baseline methods by 2.2 (+5.2\%), 1.7 (+3.6\%), 1.1 (+2.4\%), and 0.9 (+1.8\%) on MSR-VTT 
text-to-video retrieval R@1 result. For other evaluation metrics, our proposed method also outperforms the baseline method with a considerable improvement (3.0 (+6.8\%) of EMCL-Net and 3.7 (+8.0\%) of T-MASS on video-to-text R@1 result). It is worth mentioning that, based on the former SOTA method, T-MASS, adding MUSE still makes considerable improvements, especially in video-to-text retrieval. The improvement is because T-MASS mainly designed its model on the text branch and implemented fewer modifications on the video branch where our MUSE fills this blank. From this point of view, the least improvement of STAN among the four selected baselines is because that STAN has designed a new branch beside CLIP visual encoder to extract more complex video embeddings. The above results further confirm the generalizability of MUSE and the potential of MUSE to be an improvement module for any CLIP-based TVR model.\\
\noindent \textbf{Comparison with state-of-the-arts.} 
The comparisons are shown in Table \ref{tab:msrvtt_SOTA} and Table \ref{tab:other_SOTA}. As mentioned above, our proposed method is a plug-and-play model that can be applied on most ViT-based TVR models. Thus, we directly add our designed module to the former state-of-the-art (SOTA) methods and evaluate it on three popular TVR datasets. For the result of MSR-VTT, in Table \ref{tab:msrvtt_SOTA}, combined with T-MASS, our proposed method shows superior performance on almost all the evaluation metrics, achieving the new SOTA among MSR-VTT CLIP-ViT-B/32 based methods. Also we achieves SOTA performance on DiDeMo and Activity-Net by applying MUSE on the baseline models. Specifically, we select T-MASS, DiffusionRet for DiDeMo, ActivityNet considering performance and implementation difficulty. 
\subsection{Ablative Analysis of Correlation Modeling}
\begin{figure}[t]
    \centering
    \includegraphics[width=0.95\linewidth]{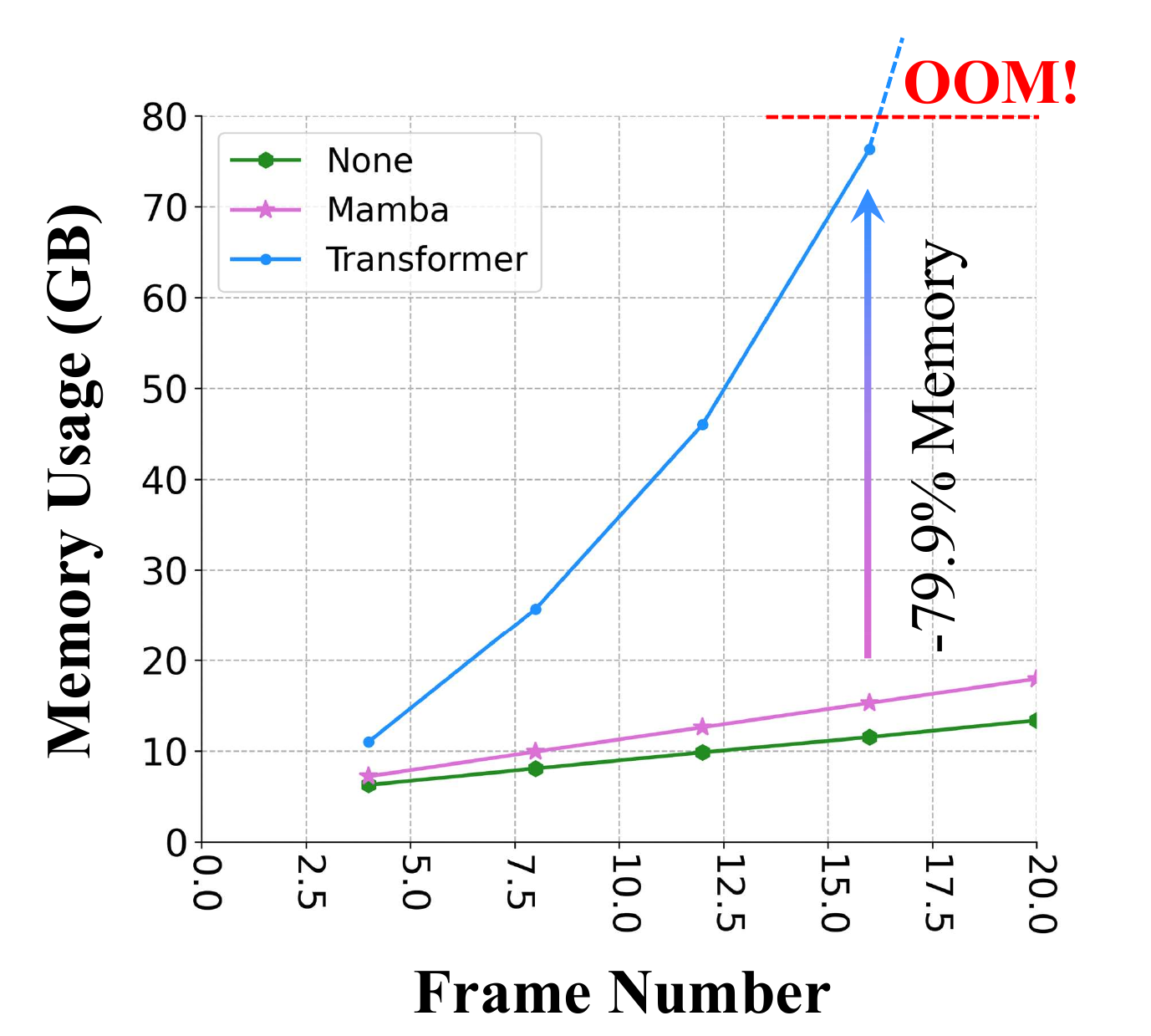}
    \caption{\textbf{Comparison of the memory usage among Transformer, Mamba, and Baseline.} The baseline selected is CLIP4clip\cite{luo2022clip4clip} with mean pooling for feature aggregation.}
    \label{fig:memory}
\end{figure}

\begin{table}[t]
\centering
\resizebox{\linewidth}{!}{
\begin{tabular}{ccccccc}
\toprule
Module & R@1↑ & R@5↑ & R@10↑ & MnR↓ & Memory(GB)↓\\
\midrule
Transformer & 43.0 & 71.1 & 80.0 & 16.3 & 36.80\\
FlashAttention & 42.6 & 69.3 & 79.7 & 16.3 & \bf 2.38\\
MambaOut & 42.4 & 70.2 & 80.7 & \bf 15.4 & 3.28\\
Mamba & \bf 44.8 & \bf 71.6 & \bf 82.1 & 15.6 & 3.40\\
\bottomrule
\end{tabular}
}
\caption{Ablations on model selection. Input frame number is set to 12. The layer number of video learner is set to 4. Memory footprint is evaluated with batch size of 16.}
\label{tab:ablate_mamba}
\end{table}
\noindent \textbf{Why using Mamba?}
To answer the question of why Mamba is our model selection, we testify the superiority of Mamba in two perspectives: \textbf{(1) Memory Usage.} We compare the memory usage between Transformer attention of quadratic complexity and Mamba of Linear complexity by calculating the memory usage as the input frame number grows. As shown in Figure \ref{fig:memory}, we test the memory usage based on CLIP4clip with batch size 16 and scale \{1, 3, 7, 14\}. The green line reflects the memory usage of the base model using mean pooling without applying MUSE. The pink line with marker star and blue line with dot reflects the memory growth when leveraging Mamba and Transformer as the video learner after multi-scale feature extraction and aggregation. From the comparison, we can easily identify that Mamba has a remarkably lower computation resource requirement than Transformer attention. For instance, when the input frame number comes to 20, the model's memory with Transformer needs more than 80GB GPU memory on a single GPU, which exceeds the memory-bound of most modern hardware accelerators. In contrast, Mamba only needs 15.37GB memory at input frame 16, which saves \textbf{79.7\%} (76.34GB) memory resources of the Transformer. Due to the linear complexity of Mamba, the memory growth is more acceptable even when it comes to 64 frames input (47.20GB in memory). \textbf{(2) Performance.} To further validate the effectiveness of Mamba architecture, we compare the performance of Mamba with Transformer and other models with linear computational complexity including MambaOut \cite{yu2024mambaout} and FlashAttention \cite{dao2022flashattention}. As shown in Table \ref{tab:ablate_mamba}, Mamba performs best on MSR-VTT among the four listed models. We replace Mamba block with other counterparts for a fair comparison. Compared with FlashAttention and Transformer, Mamba achieves better performance with almost the same memory usage as FlashAttention. To validate the effectiveness of SSM architecture in Mamba, we compare it with MambaOut which is implemented by only removing the SSM module. Results show that most of the ability to model video sequences comes from the SSM structure of Mamba.

Based on the above data, Mamba performs well in terms of performance and efficiency. Therefore, we leverage Mamba for multi-scale video sequence modeling.

\begin{table}[t]
\centering
\resizebox{0.9\linewidth}{!}{
\begin{tabular}{cccccccc}
\toprule
Scan Type & R@1↑ & R@5↑ & R@10↑ & MnR↓ \\
\midrule 
\emph{``none"} & 44.1 & 71.7 & 80.5 & \bf 14.4 \\
\emph{``v1"} & 44.0 & 71.0 & 80.7 & 14.9\\
\bf \emph{``v2"} & \bf 44.8 & \bf 71.6 & \bf 82.1 & 15.6\\
\bottomrule
\end{tabular}
}
\caption{Ablations on scan strategies.}
\label{tab:ablate_scan}
\end{table}

\begin{table}[t]
\centering
\resizebox{0.9\linewidth}{!}{
\begin{tabular}{cccccccc}
\toprule
 Agg. Mode & R@1↑ & R@5↑ & R@10↑ & MnR↓ \\
\midrule 
Frame & 43.5 & 70.6 & 80.0 & \bf 15.2\\
Spatial & 44.4 & 70.3 & 80.9 & 15.4\\
Scale & \bf 44.8 & \bf 71.6 & \bf 82.1 & 15.6\\
\bottomrule
\end{tabular}
}
\caption{Ablations on aggregation manners.}
\label{tab:ablate_agg}
\end{table}

\begin{table}[t]
\centering
\resizebox{\linewidth}{!}{
\begin{tabular}{lcccccc}
\toprule
Scale & Memory(GB)↓ & R@1↑ & R@5↑ & R@10↑ & MnR↓ \\
\midrule
\{1\} & \bf 7.62 & 43.6 & 71.2 & 81.8 & 15.2 \\
\{1, 3\}& 7.82 & 44.0  & 70.8 & 81.2 & 15.8\\
\{1, 3, 7\}& 8.76 & 44.3 & \bf 71.8 & 81.7 & 15.9 \\
\textbf{\{1, 3, 7, 14\}} & 12.60 &\bf 44.8 & 71.6 & \bf 82.1 & 15.6\\
\{1, 3, 7 14, 28\}& 29.36 & 42.5 & 71.4 & 81.6 & \bf 15.1 \\
\bottomrule
\end{tabular}
}
\caption{Ablations on video feature scale selection. \{1\} denotes the scale of \texttt{[CLS]} tokens, and numbers in curly brackets denote scale multiple times larger than \texttt{[CLS]} tokens in both width and height.}
\label{tab:ablate_scale}
\end{table}

\begin{table}[t]
\centering
\resizebox{\linewidth}{!}{
\begin{tabular}{ccccccccc}
\toprule
Layers & Memory(GB)↓ & R@1↑ & R@5↑ & R@10↑ & MnR↓ \\
\midrule
L=0 & \bf 9.20 & 42.6 & 70.8 & 79.9 & 16.1 \\
L=2 & 10.46 & 44.0 & 70.9 & 80.6 & 15.0 \\
\bf L=4 & 12.60 & 44.8 & 71.6 & \bf 82.1 & 15.6 \\
L=8 & 16.69 & 45.0 & 72.1 & 81.4 & \bf 14.6 \\
L=16 & 25.05 & \bf 45.6 & \bf 72.4 & 81.5 & 14.7\\
\bottomrule
\end{tabular}
}
\caption{Ablations on ResMamba Layer number {\emph{L}}.}
\label{tab:ablate_layer_number}
\end{table}

\subsection{Ablative Analysis of Scan Strategies}
Then, we also ablate the effects of scanning manners provided by the Mamba block from Vim \cite{zhu2024vision} in Table \ref{tab:ablate_scan}. For scan type, ``none" refers to the one-direction scan of the original Mamba block, while ``v1" and ``v2" refer to the bidirectional scan. The difference between ``v1" and 
``v2" is that ``v1" shares projection weights of forward and backward while ``v2" does not. We found that ``v2," which denotes the bidirectional scanning type with separate projection weights, performs best in video multi-scale learning. This demonstrates that scanning from low-to-high and high-to-low resolutions is crucial to the overall performance.\

\begin{figure}[t]
\includegraphics[width=1.0\linewidth]{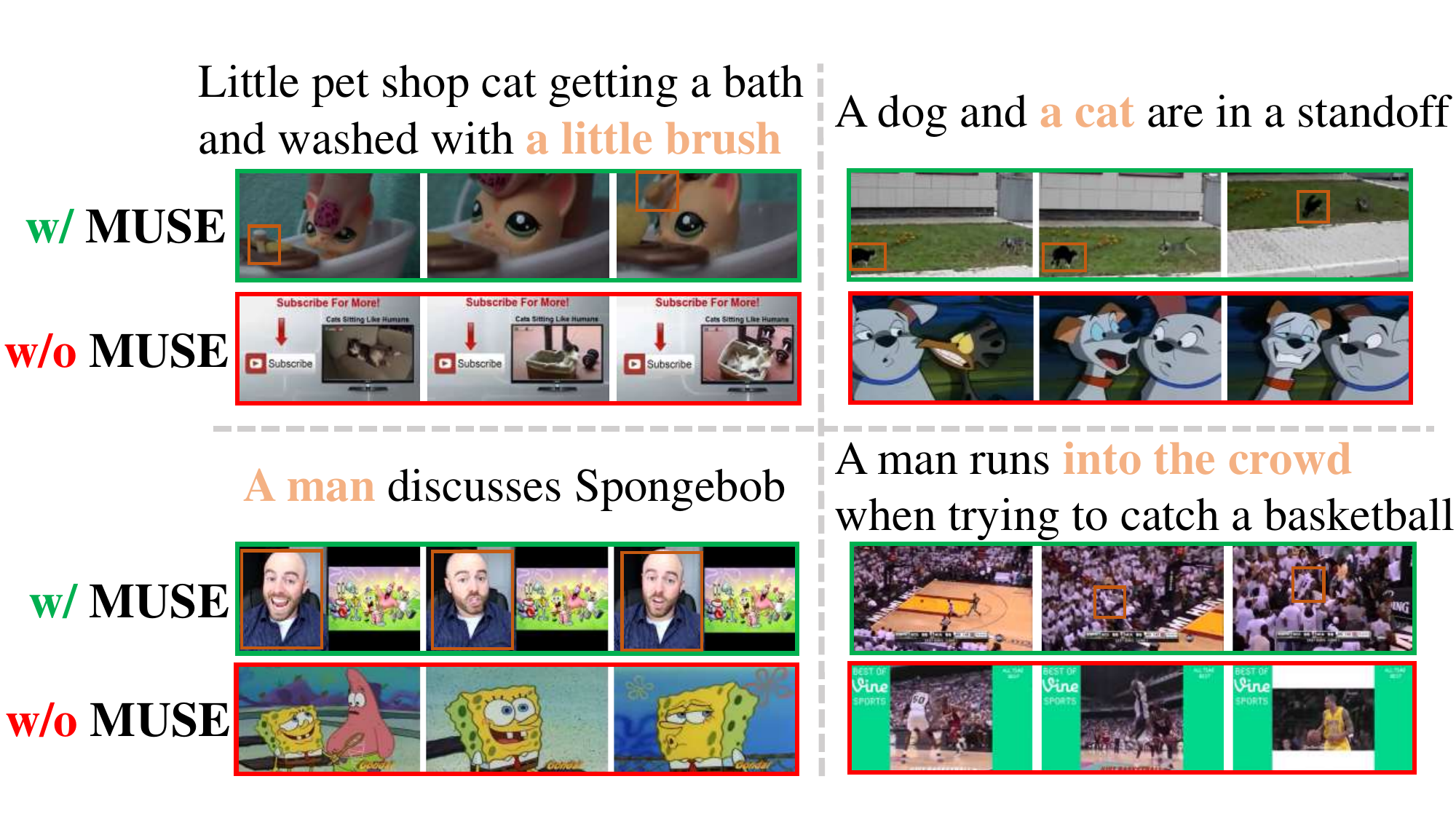}
%\vspace{-0.3cm}
\caption{\textbf{Visualization of text-video retrieval examples.} We sorted results based on their
similarity scores and visualized the rank one result. Green: correct with MUSE; Red: incorrect without MUSE. Crucial visual hints are marked with orange boxes.}
%\vspace{-0.3cm}
\label{fig:visualze}
\end{figure}

\subsection{Ablative Analysis of Scale Combination}
\noindent \textbf{Ablations on aggregation manners.}
We experiment with three different manners for feature aggregation in Figure \ref{fig:pipeline}. In Table \ref{tab:ablate_agg}, we ablate the performance of the three types of aggregation manners. We find that aggregating features scale-wise achieves the best performance for feature aggregation. This demonstrates that scale is essential for modeling video multi-scale features.\\
\noindent \textbf{Ablations on scale selection.}
Table \ref{tab:ablate_scale} compares different video scale selection manners. We find that simply adding our Mamba Learner to the original CLIP4clip at scale \{1\} still improves the performance of text-to-video retrieval results on MSR-VTT. We believe that this is the natural superiority of Mamba architecture in multi-scale video modeling. From the results, we can tell that with larger scales, our model has a trend of improving its performance except for the scale at 4x the original scale. That is because as the scale goes 4x larger, the newly added tokens get 16x longer, which brings too much redundancy when finding key information. Moreover, longer tokens also bring computation costs, which makes the model inefficient (e.g., GPU memory grows by 16.76GB). Thus, we design our method with a scale selection of \{1, 3, 7, 14\} considering both efficiency and performance.\\
\noindent \textbf{Ablations on Layer numbers.}
To further validate the effectiveness of our method, we conduct experiments on the structure of Mamba Lenarner. Specifically, we ablate the layer number $\boldsymbol{\emph{L}}$ of the ResMamba block in Table \ref{tab:ablate_layer_number}. As shown in Table \ref{tab:ablate_layer_number}, we can tell that with more ResMamba blocks, the performance still gets better, demonstrating our model design's superiority. However, we only use four layers of the ResMamba block as Mamba Learner due to the consideration of computation cost and applicability to the base models.
\subsection{Visualization Results}
In this section, we visualize some text-to-video retrieval samples from the MST-VTT testing split. As shown in Figure \ref{fig:visualze}, videos in green are the correct answers retrieved by adding MUSE on the base model and videos in red are failure results retrieved by the base model without any modification. We choose the base model as CLIP4clip with mean pooling at 12 frame input. We mark the crucial visual hints that distinguish correct answers from incorrect ones with orange boxes. The top left example reflects our model's capability of capturing small visual entity (``a little brush" in frames 1 and 3) which is essential for retrieving correct answers. Likewise, the bottom left shows that our model can notice objects of multiple granularities, e.g.,``A man" in the left part of the picture which can be treated as a large entity. This is because Mamba is an effective model for multi-scale video sequence modeling. The top right example shows that fine-grained features improve the recognition of visual entities (``a cat" rather than ``a dog" or ``a bird"). In the bottom right, the case shows our model's potential capability of identifying the relationship between visual entities (``into the crowd" is essential to distinguish the two videos). The above examples demonstrate that multi-scale features are critical for correct video retrieval and our proposed method MUSE improves this capability of the base model.

\section{Conclusion}
This paper presents MUSE acting as an efficient multi-resolution learner for text-video retrieval. Based on the plain structure of the pre-trained CLIP model, we generate multi-scale features by simply applying a feature pyramid on
the last layer feature. For the cross-resolution feature integration, we leverage Mamba to achieve effective and efficient context modeling. Extensive experiments illustrate that MUSE achieves state-of-the-art performance and scalable plug-and-play characteristics.

\section{Acknowledgments}
This work is supported in part by National Key Research and Development Program of China (2024YFE0203100) and Guangdong Provincial Key Laboratory of Ultra High Definition Immersive Media Technology (Grant No. 2024B1212010006).
\bibliography{aaai25}

% Added for Arxiv Version
\newpage
\section{Appendix}
In the appendix, we list some of the detailed illustrations of our proposed method following the suggestions of reviewers. These include illustration of multi-scale feature extraction, training time, GPU memory consumption, and illustration of ResMamba architecture.

\subsection{A. Illustration of multi-scale feature extraction}
In our implementation, we utilize several Conv2d and MaxPooling modules to extract multi-scale video features based on the output from CLIP, as discussed in the "Multi-scale Feature Generation" section. For instance, if the CLIP output has dimensions 14x14 and the scale factor is s=3, we first apply a max-pooling operation to reduce the feature map to 3x3, and then use multiple Conv2d layers with LayerNorm to extract features of this scale. Through experimentation, we have found that this straightforward approach provides the most effective results, which aligns with similar strategies discussed in the ViTDet\cite{li2022exploring} model. 

\subsection{B. Training time after adding MUSE}
Regarding the time complexity, as Table \ref{tab:training_time} shows, after incorporating MUSE, the training time on the MSR-VTT dataset increased to 17.3 GPU hours, which is a 10.9\% increase compared to the original CLIP4clip baseline of 15.6 GPU hours. This training time was measured using 12 video frames as input. Considering the significant performance improvement and the relatively low memory overhead, we believe that the slight increase in training time is acceptable for the community, especially for training text-video retrieval models.

\begin{table}[h]
\centering
\resizebox{\linewidth}{!}{
\begin{tabular}{cccccccc}
\toprule
 Method & R@1↑ & R@5↑ & R@10↑ & Time(hr)↓ \\
\midrule 
Baseline & 42.6 & 70.8 & 79.9 & \bf 15.6\\
+ MUSE & \bf 44.8 & \bf 71.6 & \bf 82.1 & 17.3\\
\bottomrule
\end{tabular}
}
\caption{Training time comparison after adding MUSE. The baseline model is CLIP4clip with 12 frame inputs.}
\label{tab:training_time}
\end{table}

\subsection{C. GPU memory consumption.}
We list the GPU memory consumption during training of the baseline model together with the models with MUSE or Transformer. As shown in Table \ref{tab:memory_consumption}, the memory required by MUSE is 12.6GB for each GPU when training with batch size 128 frame 12 on 8 A100s.
\begin{table}[h]
\centering
\resizebox{\linewidth}{!}{
\begin{tabular}{cccccccc}
\toprule
 Method & R@1↑ & R@5↑ & R@10↑ & Memory(GB)↓ \\
\midrule 
Baseline & 42.6 & 70.8 & 79.9 & \bf 9.84\\
+ MUSE & \bf 44.8 & \bf 71.6 & \bf 82.1 & 12.6 (+2.76)\\
+ Transformer & 43.0 & 71.1 & 80.0 & 36.8 (+26.96)\\
\bottomrule
\end{tabular}
}
\caption{GPU memory comparison. We list the memory consumption of the Baseline model (CLIP4clip), MUSE, and its Transformer counterpart.}
\label{tab:memory_consumption}
\end{table}
\subsection{D. Detailed illustration of ResMamba architecture}
In our paper, we advance Mamaba to ResMamba with the following two considerations: 1) Stable training: Residual connections have been widely used in various tasks for stabling gradient back-propagation; 2) Feature retaining: This simple skip-connection maintains the un-traversed features in the fused features. Specifically, we implemented a gated network for output features to pass through before being added to the input features, it consists of a LayerNorm followed by a Linear layer with zero initialization. The ablations are shown in Table \ref{tab:residual}.
\begin{table}[h]
\centering
\resizebox{\linewidth}{!}{
\begin{tabular}{cccccccc}
\toprule
 Method & R@1↑ & R@5↑ & R@10↑ & MnR↓ \\
\midrule 
w/o residual & 43.9 & \bf 71.7 & 81.6 & \bf 15.1 \\
w/ residual & \bf 44.8 & 71.6 & \bf 82.1 & 15.6 \\
\bottomrule
\end{tabular}
}
\caption{Ablations of residual architecture.}
\label{tab:residual}
\end{table}

\end{document}